\documentclass[11pt]{article}

\usepackage[final]{acl}

\usepackage{times}
\usepackage{latexsym}

\usepackage[T1]{fontenc}

\usepackage[utf8]{inputenc}
\newcommand{\rel}[1]{\texttt{#1}}

\usepackage{microtype}

\usepackage{inconsolata}

\usepackage{graphicx}

\usepackage{microtype}
\usepackage{graphicx}
\usepackage{booktabs}
\usepackage{amsmath}
\usepackage{amssymb}
\usepackage{inconsolata}

\usepackage{xcolor}
\newcommand{\zz}[1]{{\color{red} #1}}

 \usepackage[skip=3pt]{caption}
\newcommand\blfootnote[1]{%
  \begingroup\renewcommand\thefootnote{}\footnote{#1}%
  \addtocounter{footnote}{-1}\endgroup}
\title{The Instruction Tax on Factual Recall: Base Models Can Beat Their Instruction-Tuned Counterparts}

\author{
  \textbf{Christophe Khalil}\textsuperscript{*} \\
  American University of Beirut \\
  \texttt{cak29@mail.aub.edu}
  \And
  \textbf{Aya Mourad}\textsuperscript{*} \\
  Sorbonne Université, Paris, France \\
  \texttt{aya.mourad@sorbonne-universite.fr}
}

\begin{document}
\maketitle
\blfootnote{\textsuperscript{*}Equal contribution.}
\begin{abstract}
We present our solution to the AKBC LM-KBC 2026 Shared Task, where we achieved a score of 0.6836 on the official leaderboard, more than double the baseline of 0.2964. This task requires generating all possible objects for subject-relation pairs using a language model with up to 32 billion parameters, with no retrieval or fine-tuning involved. We show that the prompting methodology is as important as the model architecture: while the \emph{instruction-tuned} model version achieves 0.5681 with its chat interface, the \emph{base} model achieves 0.6836 when prompted using text completion. Thus, instruction tuning decreases factual recall; we refer to this effect as the \emph{instruction tax}, which is caused by the model's weights rather than the prompting method.Our final solution is fairly straightforward: we prompt the base model with some examples from the training data, generate 64--256 outputs, and retain only the repeated ones. Two specialized methods are then applied for each relation: one improves the recall of large answer sets, and the other verifies a person's vitality before predicting their place of death. 
\end{abstract}

\section{Introduction}Knowledge graphs (KGs) serve as an essential framework for representing factual information in a structured format, enhancing semantic search and reasoning. Manual curation of such frameworks is expensive, and the resulting graphs always contain substantial information gaps \cite{schneider2022decade, hofer2024construction}. Although large language models (LLMs) have learned tremendous amounts of factual knowledge implicitly during their pre-training \cite{petroni2019lama}, extracting this knowledge is challenging due to prompt sensitivity and hallucinations \cite{jiang2020can, zhang2025siren}.

To evaluate progress on these extraction challenges, the AKBC 2026 Shared Task\footnote{\url{https://lm-kbc.github.io/challenge2026/}} assesses open-weight LLMs with fewer than 32 billion parameters under a strict \emph{closed-book} setting, where retrieval, external knowledge sources, and fine-tuning are prohibited. The task requires models to predict all valid objects for a given subject--relation pair. For instance, the query \emph{(Albert Einstein, city-of-death)} should produce the answer \emph{{Princeton}}, whereas the query \emph{(Barack Obama, city-of-death)} should return an empty set because no valid answer exists. This ability to abstain from producing unsupported facts is an essential part of the benchmark, as nearly 20\% of the evaluation instances cannot be answered.

Our system is inspired by an empirical finding that we name the \emph{instruction tax}. While \emph{instruction-tuned} models \citep{ouyang2022instruct} are generally favored for question answering tasks, our findings show that \emph{instruction-tuned} models can sometimes significantly under-perform \emph{base} models in retrieving facts. For instance, on the area relation, the \emph{base} model queried by simple text completion outperforms the \emph{instruction-tuned} model version by 0.29 through its chat interface, and by 0.30 under the same completion prompt. This gap persists despite prompting the \emph{instruction-tuned} models using completion-style prompts.

Building on this finding, we developed a system based on five methodological principles: (1) choosing the \emph{base}  model through a fast probing process that takes about one GPU-hour; (2) prompting the \emph{base}  model using few-shot continuations rather than asking questions explicitly; (3) sampling 64--256 times for each row, and keeping only what recurs and returning empty if there is no recurrence; (4) adding a verification step for low-frequency examples in relations with hundreds of possible answers; and (5) breaking down more complicated tasks, such as checking whether the person is dead before predicting the \emph{city-of-death}. Code, prompts, fitted thresholds, and the commands that reproduce each table are available at \url{https://github.com/aiamourad/lm-kbc-2026/}.

\section{Related Work}Language models have been tested for factual recall in multiple studies. The LAMA benchmark was introduced by \citet{petroni2019lama} and consists of factual questions answered using the fill-in-the-blank format. \citet{roberts2020closedbook} showed that a fine-tuned model can answer factual questions from its parameters alone, and \citet{brown2020fewshot} showed the same without weight updates, using
a few in-context examples as prompts. Still, factual recall is notably less effective for long-tail entities \citep{sun2024headtotail}, making long-tail entities the hard case for our task. The LM-KBC benchmark \citep{singhania2022lmkbc} adds an evaluation setting for factual recall that requires predicting the entire set of answers and compares the predictions against accepted aliases.

Instruction tuning enhances the ability to follow instructions but can impair other abilities \citep{ouyang2022instruct}, with \citet{zhou2023lima} arguing that it alters behavior more than knowledge. This paper quantifies these impacts in terms of factual recall. In addition, our method employs repeated sampling, akin to the self-consistency approach \citep{wang2023selfconsistency, manakul2023selfcheckgpt}, and examines the trade-off between efficiency and recall in mixture-of-experts models \citep{fedus2022switch}.

\section{AKBC Shared Task \& Metrics}\subsection{Dataset}
This task contains six relationships and has 477, 475, and 475 examples in the train/val/test splits, respectively. There are two numerical relationships (\texttt{hasArea}, \texttt{hasCapacity}) for which the answer is a single number, and it is correct if it differs from the ground-truth answer by no more than ±5\%. \\zz{\texttt{awardWonBy} may have up to several hundred objects that the model must list to be considered correct. The remaining three relationships (countries (or comparable territories) that share a land border with the subject (\texttt{countryLandBordersCountry}), the stock exchange(s) on which the company's shares are publicly traded (\texttt{companyTradesAtStockExchange}), the city where the person died (\texttt{personHasCityOfDeath})) have small answer sets that are empty for 18--42\% of training rows.}

\subsection{Official Scorer}
The official scorer normalizes strings, maps a prediction to an entity if it equals one of that entity's aliases, and maps predictions to entities to maximize the number of matches; row scoring is done independently, and the scores are then averaged. Let $S$ denote the predicted set, $G$ the reference set, and $T$ the number of matches. The row score is:
\begin{equation}
\label{eq:rowf1}
F_1(S,G) =
\begin{cases}
1 & S = G = \emptyset\\
0 & \text{one of S,G \text{ empty}}\\
2T/(|S|{+}|G|) & \text{otherwise.}
\end{cases}
\end{equation}
The first two cases derive from the scoring conventions: abstention on an empty-answer row scores 1.0, whereas guessing on an empty-answer row scores 0. The baseline provided by the organizers never abstains and hence scores 0 on every empty-answer row. The leaderboard uses this scorer, and our validation scores come from running it locally.
\section{Model Selection}\paragraph{\emph{The knowledge probe}}

Which model should we select, and how should we interact with it? To this end, we provide the answers by means of an easy and cheap procedure called the \emph{knowledge probe}. It consists of about 100 examples with known answers. For each of them, we query each of the models under consideration using several phrasings to ask for the corresponding fact and sample their answers 8--12 times. Next, we compute how often each particular answer was generated.

This results in two scores for each candidate model. The first one is called \emph{voted} and stands for the proportion of examples whose most frequently returned answer is the correct one. The second one is the \emph{ceiling}, i.e., the proportion of examples for which the correct answer appears at least once in the generated responses. This score reflects the upper limit of performance under such a probing procedure, since no selection strategy can return an answer that does not occur in the samples.

We probe the two numeric relations, as each sample yields a single number that can be parsed and checked automatically against the 5\% tolerance threshold.

\begin{table}[t]
\centering\footnotesize
\setlength{\tabcolsep}{0.7pt}
\begin{tabular}{lccccc}
\toprule
& \multicolumn{3}{c}{\emph{voted} accuracy} & tax & \emph{ceiling} \\
\cmidrule(lr){2-4}\cmidrule(lr){5-5}\cmidrule(lr){6-6}
& base & \multicolumn{2}{c}{it-tuned} & & \\
\cmidrule(lr){2-2}\cmidrule(lr){3-4}
family & compl. & chat & compl. & $\Delta$ & \shortstack{base\\$\to$chat} \\
\midrule
\multicolumn{6}{l}{\emph{Families publishing both model versions}} \\
gemma-3-12b & .720 &.370 & .320 & $+$.350 & .84$\to$.47 \\
gemma-3-27b & \textbf{.880} & .590 & .580 & $+$.290 & .95$\to$.69 \\
gemma-2-27b & .720 & .640 & .640 & $+$.080 & .82$\to$.77 \\
Mistral-Nemo-2407 & .710 & .630 & .660 & $+$.080 & .90$\to$.87 \\
\midrule
\multicolumn{6}{l}{\emph{Instruction-tuned only}} \\
Mistral-Small-3.2-24B & --- & .720 & .690 & --- & --- \\
Magistral-Small-2506 & --- & .610 & .570 & --- & --- \\
Qwen3-30B-A3B (MoE) & --- & .430 & .430 & --- & --- \\
Qwen3-30B-A3B-Think. & --- & .430 & .410 & --- & --- \\
\bottomrule
\end{tabular}
\caption{Knowledge probe on \rel{hasArea}, 100 validation rows, leading zeros
omitted. Each row is one weights family, queried three ways: the
base model by completion, and the it-tuned model by chat and by completion.
\emph{voted} is the accuracy of the best question form. $\Delta$ is the probe
tax, base minus chat, on this relation alone; the full-system tax is
Table~\ref{tab:models}. \emph{ceiling} is the share of rows whose correct answer
appears in at least one sample, base against chat.}
\label{tab:probe}
\end{table}

\paragraph{Which model version to ship}

Table~\ref{tab:probe} highlights the results that were derived from around one GPU-hour of computing effort per model version, and this guided the selection of the model. In the probe we ask for the same fact in several different wordings, such as "The area of $X$ is" or "$X$ covers an area of". For each model version, we then report only the score of its best wording. We pick that wording using the same 100 rows we score on, so these numbers are a little too high. We treat every model version in the same way, but this helps a model more when its wordings give very different results. The table is therefore best used to rank the model versions, not to predict what one fixed prompt would score. Dense models outperformed the mixture-of-experts (MoE) model across all the experiments. While the evaluation counts an MoE model by its total number of parameters, not all of these parameters get activated per token, and the performance appears to depend on the amount of such computation. The reasoning-tuned model does not provide any noticeable improvement.

\paragraph{\emph{The instruction tax}}

We observe a discrepancy between the scores of the base 0.880 and instruction-tuned 0.59 version of the gemma-3-27b model. To analyze whether the following is resulting from the nature of the prompt, we prompt the instruction-tuned model using the same completion prompt (compl. under it-tuned in Table \ref{tab:probe}) used for the base model version. Even with an identical query, it fell short at 0.580, and the sample ceiling falls from 0.950 to 0.690. This gives us a clearer image that the tuned model forgot facts it learned.

To check how consistent this pattern is, we test three additional pairs (gemma-3-12b, gemma-2-27b, and Mistral-Nemo-2407). In all eight tests, the base model version is at least as good as the instruction-tuned model version on both numerical relations. This appears to be a feature not inherent to instruction tuning per se, but dependent on the particular tuning run. Inside the Gemma family, the difference is 0.290 for gemma-3-27b and 0.350 for gemma-3-12b, but only 0.080 for gemma-2-27b, even though it is the same size as gemma-3-27b. In other words, with one generation less, the same tuning carries only a quarter of the cost. Mistral-Nemo-2407 also shows this small difference of 0.080.

The key difference between the two cases lies in the ceiling. With a high tax, performance on the matched completion prompt becomes very poor: dropping from 0.950 to 0.690 in gemma-3-27b, and from 0.840 to 0.470 in gemma-3-12b. With a low tax, the ceiling remains almost intact, dropping merely from 0.820 to 0.770 in gemma-2-27b, and from 0.900 to 0.870 in Mistral-Nemo-2407. It seems, therefore, that a high tax results in the loss of some answers from the sampling distribution, whereas a low tax results in poor answer selection from the remaining sample. Since both can be quantified using the same one-hour probe, the distinction can be made prior to committing to a model version.

As shown in Section~\ref{sec:results}, the probe order matches the order of the official submission runs, suggesting that the probing evaluation serves as an effective proxy for full submissions. The final system employs the base model version, gemma-3-27b-pt \citep{gemmateam2025gemma3}, for all six relations.
\begin{table*}[h]
\centering
\footnotesize
\setlength{\tabcolsep}{5pt}
\begin{tabular}{lp{93mm}cc}
\toprule
Relation & Completion format & $k$ & $n$ \\
\midrule
\rel{hasArea}
&
The total area of \{s\} is \{v\} square kilometres.
& 5 & 64 \\

\rel{hasCapacity}
&
The seating capacity of \{s\} is \{v\}.
& 5 & 64 \\

\rel{countryLandBordersCountry}
&
Countries sharing a land border with \{s\}: \{v\}
& 12 & 128 \\

\rel{companyTradesAtStockExchange}
&
\{s\} stock exchange listings: \{v\}
& 12 & 256 \\

\rel{personHasCityOfDeath}
&
City where \{s\} died (or \texttt{none} if alive as of 1 July 2026):
\{v\}
& 12 & 128 \\

\rel{awardWonBy}
&
Recipients of \{s\}: \{v\}
& 4 & 64 \\
\bottomrule
\end{tabular}
\caption{Completion formats and sampling budgets ($k$: exemplars, $n$: samples
per row); \{s\} is the subject and \{v\} the gold value or \texttt{none}.}
\label{tab:formats}
\end{table*}
\section{Methodology}
\label{sec:system}
For inference, we deploy only one model (\textit{gemma-3-27b-pt}) for the whole set of tasks. This model has 27.43 billion parameters, obtained precisely by counting the released weights rather than relying on the model's name, which may be incorrect. Within the task's 32-billion-parameter budget, it is impossible to deploy any other useful model in parallel with it. Inference is provided with the help of vLLM \citep{kwon2023vllm} using two H100 GPUs and thus is performed for the whole test split in just a few minutes.

\paragraph{Never ask a question.} For each relation, there is one completion template (see Table~\ref{tab:formats}). A prompt is composed of a few examples selected from the training dataset and ends with an unfinished example that corresponds to the subject of the query:

\begin{quote}\small\ttfamily
The total area of Lake Victoria is 68800 square kilometres.\\
The total area of Bodensee is 536 square kilometres.\\
The total area of Vygozero is
\end{quote}

\noindent Such a formulation will naturally encourage the continuation of the observed pattern rather than an interpretation of the input as an instruction. There is no chat template and no task-specific instruction, and the model receives only the text in the exact form it should produce.

There are two crucial design decisions here. First of all, the numerical examples are chosen in such a way that they cover the whole distribution of numbers in the training dataset, including one sample from each quintile. This will prevent a bias towards a certain numerical range in the prompt. Second, examples with an empty response in the case of set-valued relations make up about one third of the exemplars and are marked as "none".

\paragraph{Ask many times, keep what repeats.} Each row is sampled $n$ times with different random seeds, and every output is truncated at the first line break. The temperature is 0.7 throughout. The exact value matters little: between 0.3 and 1.3 the differences stay within noise once the decision thresholds are refitted. Only very low temperatures do real damage, making the 64 samples nearly identical, collapsing the process to a single completion and removing the small benefit of aggregation.

We then parse the outputs into the required format for each relation type: a single numeric answer for numeric relations, and a comma-separated list for set-valued relations, where the output "none" denotes the empty set. Prior to aggregation, the responses are normalized in the same way as in the evaluation script: string-valued answers are merged according to the scorer's normalization rules, and numeric answers are grouped using the scorer's 5\% tolerance.

For numerical relations, denote the parsed samples as $v_1, \dots, v_m$, where $m \le n$ because not every output parses into a number. For each candidate $c$, the cluster $C(c)$ is formed by all those samples lying within 5\% of $c$, and the prediction is the median of the biggest cluster:
\begin{equation}
\label{eq:numeric}
\hat{v} = \operatorname{median} C(c^{\ast}), \quad
c^{\ast} = \operatorname*{arg\,max}_{c} |C(c)|,
\end{equation}
where $C(c) = \{\, v_i : |v_i - c| \le 0.05\,c \,\}$. This can be considered as majority voting modified to match the evaluation function: since the scorer allows any prediction that is within 5 percent of the reference value, votes are counted on tolerance clusters and not on numeric equality.

For relations with sets, we define $A_1, \dots, A_m$ as the sets of answers after parsing. Support $s(a)$ for a candidate answer $a$ is defined as its frequency of recurrence, that is, the fraction of the $A_j$ that contain $a$:
\begin{equation}
\label{eq:support}
s(a) = \tfrac{1}{m}\bigl|\{ j : a \in A_j \}\bigr|,
\quad
S = \{ a : s(a) \ge \tau_r \}.
\end{equation}

The threshold $\tau_r$ is the sole parameter fitted for each relation independently. Instead of picking it manually, we pick the one that achieves the best official score on the training examples, and re-fit it each time either the number of samples $n$ or the number of exemplars $k$ is modified. These thresholds are more conservative for those relations where misprediction carries a high cost. In particular, for \rel{personHasCityOfDeath}, the system will predict the city only if at least 40\% of the 128 samples vote for it. For those relations having a bigger reference set, the fitted threshold is typically smaller. Abstention is a natural consequence of the same approach, since if none of the candidates achieves the needed recurrence rate, the predicted set is empty.

\paragraph{Verifying rare candidates for \rel{awardWonBy}.} The relation \rel{awardWonBy} is especially challenging because its answer set can have hundreds of names. In such a situation, it is difficult to differentiate the genuine candidate, which only occurs three times among 64 samples, from the hallucinated candidate by recurrence alone. For this reason, we employ the model as a second-stage verifier using completion. We first present the model with ten verification examples from the training set that are evenly divided into positive and negative examples (e.g., \texttt{Did Barack Obama get the Nobel Peace Prize?~yes} and \texttt{Did Marie Curie get the Turing Award?~no}), and then pose the same question to the model regarding the candidate. We sample the verification question eight times, and let $q(a)$ be the ratio of samples answered with yes. We then add back the candidates whose support is at least $2/64$ and whose $q(a) \ge 0.75$.
The two threshold values are chosen based on training examples. In the verification process, true positives are separated from false positives by means of the Area Under the Curve (AUC), which has a value of 0.86 on training candidates (0.75 on validation candidates). Lastly, we perform a non-neural clean-up procedure to combine duplicate names such as "Robert Caro" and "Robert A.\ Caro". Without performing this step, the scorer would consider both entries to be different predictions.

\paragraph{Is the person dead or alive?} Approximately 40\% of the death-relation rows refer to people who are still alive, and the correct answer in such cases is the empty set. We took a two-step approach in which we identify whether a given subject is dead or alive. If the person is alive, we ask the model to assign the empty set. On the other hand, if the model is confident that the person is indeed dead, we ask it to recall the location of their death.

Information about whether the person is alive or dead is far more prevalent than information about where they died. The location of the death is tail knowledge, while the person's living status is more widespread. We exploit this difference as an initial filter, which we call the \emph{existence check}: it predicts whether an object exists for the row at all, rather than which object it is. For every query, we take 16 samples of a separate completion, built from ten exemplars evenly split between living and dead subjects of the training split, of the form \texttt{As of 1 July 2026, \{s\} is alive/dead}, followed by the query subject. If all 16 samples predict that the person is alive, then the answer must be the empty set.

\section{Results}\label{sec:results}
Table~\ref{tab:main} gives the final scores and Table~\ref{tab:models} the
comparisons across models and interfaces, all from official submissions.
Relations contribute 100 rows each except \rel{awardWonBy} (10),
\rel{countryLandBordersCountry} (67) so
"All relations" averages over 475 rows, not six scores; the last row scores
the abstention decision, with the empty set as the positive class.
\begin{table}[h]
\centering\footnotesize
\setlength{\tabcolsep}{1.5pt}
\begin{tabular}{lcccc}
\toprule
 & val & \multicolumn{3}{c}{test (official)} \\
\cmidrule(lr){3-5}
Relation & F1 & P & R & F1 \\
\midrule
\texttt{awardWonBy} & 0.266 & 0.389 & 0.343 & 0.338 \\
\texttt{companyTradesAtStockExchange} & 0.765 & 0.893 & 0.932 & 0.872 \\
\texttt{countryLandBordersCountry} & 0.989 & 0.962 & 0.976 & 0.958 \\
\texttt{hasArea} & 0.900 & 0.820 & 0.820 & 0.820 \\
\texttt{hasCapacity} & 0.299 & 0.316 & 0.316 & 0.316 \\
\texttt{personHasCityOfDeath} & 0.550 & 0.900 & 0.630 & 0.570 \\
\midrule
\textbf{All relations} & \textbf{0.669} & \textbf{0.759} & \textbf{0.712}
  & \textbf{0.684} \\
Abstention decision & --- & .722 & 0.892 & 0.798 \\
\bottomrule
\end{tabular}
\caption{Final system. Test columns are official; validation is the same scorer
run locally. Formats and thresholds are fitted on training rows only.}
\label{tab:main}
\end{table}

Firstly, there is an \emph{instruction tax}, which is significant, especially when it comes to relations that rely heavily on exact recall (Table~\ref{tab:models}). Within one weights family, using the \emph{base}  model for completion instead of the instruction-tuned model via the chat interface increases the score by $+0.116$. The greatest improvement can be seen in the numeric relations, namely area ({$+0.203$}) and capacity ($+0.156$). In fact, the \emph{base}  model outperforms the instruction-tuned model on every single relation.

\begin{table}[t]
\centering\footnotesize
\setlength{\tabcolsep}{0.78pt}
\begin{tabular}{lccc}
\toprule
 & \multicolumn{3}{c}{official test F1} \\
\cmidrule(lr){2-4}
Relation & it, chat & pt, compl. & $\Delta$ \\
\midrule
\texttt{hasArea} & 0.617 & 0.820 & $+$0.203 \\
\texttt{hasCapacity} & 0.160 & 0.316 & $+$0.156 \\
\texttt{companyTradesAtStockExchange} & 0.784 & 0.872 & $+$0.088 \\
\texttt{personHasCityOfDeath} & 0.480 & 0.570 & $+$0.090 \\
\texttt{awardWonBy} & 0.271 & 0.338 & $+$0.067 \\
\texttt{countryLandBordersCountry} & 0.946 & 0.958 & $+$0.012 \\
\midrule
\textbf{All relations} & 0.568 & \textbf{0.684} & $+$0.116 \\
\midrule
Mistral-Small-3.2-24B, chat & \multicolumn{3}{c}{0.613} \\
Mistral-Small-3.2-24B, compl. & \multicolumn{3}{c}{0.615} \\
Magistral-Small-2506, compl. & \multicolumn{3}{c}{0.578} \\
Qwen3-30B-A3B (MoE), chat & \multicolumn{3}{c}{0.534} \\
Qwen3-30B-A3B (MoE), compl. & \multicolumn{3}{c}{0.551} \\
Qwen3-30B-A3B-Think., compl. & \multicolumn{3}{c}{0.510} \\
task baseline & \multicolumn{3}{c}{0.2964} \\
\bottomrule
\end{tabular}
\caption{The \emph{instruction tax} at full-system scale, all scores official.
Upper block: the gemma-3-27b family, gemma-3-27b-it by chat
against gemma-3-27b-pt by completion, with $\Delta$ the second minus the first
for each relation. Lower block: model versions with no base
sibling, so there is nothing to subtract; each row is a single score, marked
\emph{chat} when the model was asked through its chat interface and
\emph{compl.} when it was given the completion prompt instead.}
\label{tab:models}
\end{table}
The comparison only makes sense if the instruction-based model is a strong one. To ensure this, we created a detailed prompt with a few-shot examples and challenged the model with yes-or-no questions for each query. More details are provided in Appendix~\ref{app:chat}. Even after giving the instruction-tuned system all these additional advantages, the simpler completion-based system still performed better on every relation. Our findings imply a strong difference between the \emph{base} and \emph{instruction-tuned} models when it comes to factual knowledge recall: the facts recoverable by the \emph{base} model become extremely rare in the output of the \emph{instruction-tuned} model.

Secondly, the ranking method of Table~\ref{tab:probe} has predicted each ordering in the official submission runs accurately, as the lower block of Table~\ref{tab:models} shows. The dense models were ranked ahead of the MoE model (0.568 and 0.613 against 0.534) under the same instruction-tuned system, and the \emph{base}  checkpoint was ranked ahead of all of them at 0.681. The point is that the decisions which are key to achieving the highest score, model and interface, can be determined using only one GPU hour of computing time.
Thirdly, sampling should be increased until little or nothing is gained, and this appears to be contingent on the relation. The official score increased from 0.6218 with $n{=}8$ samples to 0.6292 with $n{=}24$ samples, and again to 0.6604 with $n{=}64$ samples, the latter one being computed using the company format in Table~\ref{tab:formats}. More samples were useful for some relations; for set relations, the gains continued until $n{=}128$, while for the company relation, the optimal number was found to be $n{=}256$. The verification rescue contributed an additional $+0.027$ on its relation, and the existence check contributed $+0.030$ on the death relation, which gave 0.6809. Doubling the samples for the numeric relations ($n{=}128$) did not help, so $n{=}64$ already saturates them. 

Last, much of the gain comes from knowing when to say nothing (last two rows of Table~\ref{tab:main}). Empty exemplar lines plus the training-fitted threshold give an abstention F1 of 0.773; the existence check raises it to 0.798 and death precision to 0.900. The check catches an error recurrence cannot: the model naming a city, consistently, for someone still alive.

\section{Conclusion}The design of our system rests on one insight: when it comes to factual recall, instruction tuning is not worth the expense. On the same parameters, gemma-3-27b-it scores 0.568 through its chat interface, while gemma-3-27b-pt scores 0.684 by completion. The rest follows: a one-GPU-hour probe selects the model version and the interface, each row is sampled between 64 and 256 times from the completion prompt, and only those responses above a fitted threshold are retained, making abstention the norm rather than the exception. The final score is 0.6836, against 0.2964 for the task’s baseline. Two takeaways should apply elsewhere: the model version and the interface need to be tested, and where the best answer is often empty, knowing when not to answer helps almost as much as knowing more.

\section*{Limitations}

The instruction tax is calculated on four (base, instruct) pairs. Officially, changing the weights alone costs 0.085 for gemma-3-27b and gemma-3-12b, whose sampling ceiling collapses, but 0.016 for Mistral-Nemo-2407 and 0.000 for gemma-2-27b, whose ceiling holds: the size of the tax depends on the tuning process, and four families are not enough to determine its applicability. The system is also quite costly, taking 64--256 generations to process each row, while the instruction-tuned baseline uses only one, and it is not zero-shot: the prompts are constructed from the training rows with labels, and the thresholds have to be recalibrated every time either $n$ or $k$ changes. The verification rescue and the existence check were each crafted by hand for a single relation. 

\section*{Acknowledgments}
This work was performed using HPC resources from GENCI-IDRIS (Grant 2025-AD011016170).

\bibliography{custom}

@inproceedings{petroni2019lama,
  title     = {Language Models as Knowledge Bases?},
  author    = {Petroni, Fabio and Rockt{\"a}schel, Tim and Riedel, Sebastian and
               Lewis, Patrick and Bakhtin, Anton and Wu, Yuxiang and Miller, Alexander},
  booktitle = {Proceedings of the 2019 Conference on Empirical Methods in Natural
               Language Processing and the 9th International Joint Conference on
               Natural Language Processing (EMNLP-IJCNLP)},
  pages     = {2463--2473},
  year      = {2019}
}

@inproceedings{wang2023selfconsistency,
  title     = {Self-Consistency Improves Chain of Thought Reasoning in Language Models},
  author    = {Wang, Xuezhi and Wei, Jason and Schuurmans, Dale and Le, Quoc V. and
               Chi, Ed H. and Narang, Sharan and Chowdhery, Aakanksha and Zhou, Denny},
  booktitle = {The Eleventh International Conference on Learning Representations (ICLR)},
  year      = {2023}
}

@inproceedings{kwon2023vllm,
  title     = {Efficient Memory Management for Large Language Model Serving with PagedAttention},
  author    = {Kwon, Woosuk and Li, Zhuohan and Zhuang, Siyuan and Sheng, Ying and
               Zheng, Lianmin and Yu, Cody Hao and Gonzalez, Joseph E. and
               Zhang, Hao and Stoica, Ion},
  booktitle = {Proceedings of the 29th Symposium on Operating Systems Principles (SOSP)},
  year      = {2023}
}

@inproceedings{dembczynski2011exact,
  title     = {An Exact Algorithm for F-Measure Maximization},
  author    = {Dembczy{\'n}ski, Krzysztof and Waegeman, Willem and Cheng, Weiwei and
               H{\"u}llermeier, Eyke},
  booktitle = {Advances in Neural Information Processing Systems (NIPS)},
  year      = {2011}
}

@inproceedings{ye2012optimizing,
  title     = {Optimizing F-Measures: A Tale of Two Approaches},
  author    = {Ye, Nan and Chai, Kian Ming A. and Lee, Wee Sun and Chieu, Hai Leong},
  booktitle = {Proceedings of the 29th International Conference on Machine Learning (ICML)},
  year      = {2012}
}

@inproceedings{roberts2020closedbook,
  author    = {Adam Roberts and Colin Raffel and Noam Shazeer},
  title     = {How Much Knowledge Can You Pack Into the Parameters of a Language Model?},
  booktitle = {Proceedings of the 2020 Conference on Empirical Methods in Natural
               Language Processing (EMNLP)},
  year      = {2020}
}

@inproceedings{singhania2022lmkbc,
  author    = {Sneha Singhania and Tuan-Phong Nguyen and Simon Razniewski},
  title     = {{LM-KBC}: Knowledge Base Construction from Pre-trained Language Models},
  booktitle = {Proceedings of the Semantic Web Challenge on Knowledge Base
               Construction from Pre-trained Language Models ({LM-KBC} @ {ISWC})},
  series    = {CEUR Workshop Proceedings},
  year      = {2022}
}

@inproceedings{manakul2023selfcheckgpt,
  author    = {Potsawee Manakul and Adian Liusie and Mark J. F. Gales},
  title     = {{SelfCheckGPT}: Zero-Resource Black-Box Hallucination Detection
               for Generative Large Language Models},
  booktitle = {Proceedings of the 2023 Conference on Empirical Methods in Natural
               Language Processing (EMNLP)},
  year      = {2023}
}

@article{fedus2022switch,
  author  = {William Fedus and Barret Zoph and Noam Shazeer},
  title   = {Switch Transformers: Scaling to Trillion Parameter Models with
             Simple and Efficient Sparsity},
  journal = {Journal of Machine Learning Research},
  volume  = {23},
  year    = {2022}
}

@article{gemmateam2025gemma3,
  author  = {{Gemma Team}},
  title   = {Gemma 3 Technical Report},
  journal = {arXiv preprint arXiv:2503.19786},
  year    = {2025}
}

@inproceedings{sun2024headtotail,
  author    = {Kai Sun and Yifan Ethan Xu and Hanwen Zha and Yue Liu and Xin Luna Dong},
  title     = {Head-to-Tail: How Knowledgeable are Large Language Models ({LLM}s)?
               {A.K.A.} Will {LLM}s Replace Knowledge Graphs?},
  booktitle = {Proceedings of the 2024 Conference of the North American Chapter of
               the Association for Computational Linguistics (NAACL)},
  year      = {2024}
}

@inproceedings{brown2020fewshot,
  title = {Language Models are Few-Shot Learners},
  author = {Brown, Tom and Mann, Benjamin and Ryder, Nick and Subbiah, Melanie and Kaplan, Jared and Dhariwal, Prafulla and Neelakantan, Arvind and Shyam, Pranav and Sastry, Girish and Askell, Amanda and others},
  booktitle = {Advances in Neural Information Processing Systems},
  volume = {33},
  year = {2020}
}

@inproceedings{ouyang2022instruct,
  title = {Training Language Models to Follow Instructions with Human Feedback},
  author = {Ouyang, Long and Wu, Jeffrey and Jiang, Xu and Almeida, Diogo and Wainwright, Carroll and Mishkin, Pamela and Zhang, Chong and Agarwal, Sandhini and Slama, Katarina and Ray, Alex and others},
  booktitle = {Advances in Neural Information Processing Systems},
  volume = {35},
  year = {2022}
}

@inproceedings{zhou2023lima,
  title = {{LIMA}: Less Is More for Alignment},
  author = {Zhou, Chunting and Liu, Pengfei and Xu, Puxin and Iyer, Srini and Sun, Jiao and Mao, Yuning and Ma, Xuezhe and Efrat, Avia and Yu, Ping and Yu, Lili and others},
  booktitle = {Advances in Neural Information Processing Systems},
  volume = {36},
  year = {2023}
}

@inproceedings{schneider2022decade,
  title={A decade of knowledge graphs in natural language processing: A survey},
  author={Schneider, Phillip and Schopf, Tim and Vladika, Juraj and Galkin, Mikhail and Simperl, Elena and Matthes, Florian},
  booktitle={Proceedings of the 2nd Conference of the Asia-Pacific Chapter of the Association for Computational Linguistics and the 12th International Joint Conference on Natural Language Processing (Volume 1: Long Papers)},
  pages={601--614},
  year={2022}
}

@article{hofer2024construction,
  title={Construction of knowledge graphs: Current state and challenges},
  author={Hofer, Marvin and Obraczka, Daniel and Saeedi, Alieh and K{\"o}pcke, Hanna and Rahm, Erhard},
  journal={Information},
  volume={15},
  number={8},
  pages={509},
  year={2024},
  publisher={MDPI}
}

@article{jiang2020can,
  title={How can we know what language models know?},
  author={Jiang, Zhengbao and Xu, Frank F and Araki, Jun and Neubig, Graham},
  journal={Transactions of the Association for Computational Linguistics},
  volume={8},
  pages={423--438},
  year={2020},
  publisher={MIT Press}
}

@article{zhang2025siren,
  title={Siren's Song in the {AI} Ocean: A Survey on Hallucination in Large Language Models},
  author={Zhang, Yue and Li, Yafu and Cui, Leyang and Cai, Deng and Liu, Lemao and Fu, Tingchen and Huang, Xinting and Zhao, Enbo and Zhang, Yu and Chen, Yulong and others},
  journal={Computational Linguistics},
  volume={51},
  number={4},
  pages={1373--1418},
  year={2025},
  publisher={MIT Press}
}

\appendix

\section{Base and Instruction-Tuned Checkpoints, End to End}
\label{app:pairs}
In Table~\ref{tab:pairs}, the whole system developed in Section~\ref{sec:system} is run
on each family releasing both checkpoints, with one call per checkpoint, and
is evaluated on the official test set. Everything else in the pipeline, the
sampling budget, and the training process is exactly the same for the eight
runs; the only difference is in the weights. The probe columns reproduce
Table~\ref{tab:probe}. Eight runs cost 7 GPU-hours on two H100 GPUs.

\begin{table}[h]
\centering\footnotesize
\setlength{\tabcolsep}{1.5pt}
\begin{tabular}{lcccccc}
\toprule
& \multicolumn{3}{c}{official test} & \multicolumn{2}{c}{probe} \\
\cmidrule(lr){2-4}\cmidrule(lr){5-6}
family & base & tuned & $\Delta$ & $\Delta$ & ceiling \\
\midrule
gemma-3-27b & .6836 & .5932 & $+$.0904 & .290 & .95$\to$.69 \\
gemma-3-12b & .6018 & .5166 & $+$.085 & .350 & .84$\to$.47 \\
Mistral-Nemo & .5881 & .5724 & $+$.016 & .080 & .90$\to$.87 \\
gemma-2-27b & .6190 & .6189 & $+$.000 & .080 & .82$\to$.77 \\
\bottomrule
\end{tabular}
\caption{The complete system on four base/instruct pairs, official test, leading
zeros omitted. $\Delta$: cost of the tuned weights; \emph{ceiling} from
Table~\ref{tab:probe}.}
\label{tab:pairs}
\end{table}

The cost of each of these jobs is given in Table~\ref{tab:compute}. All jobs use two H100 GPUs, and thus the number of GPU hours is double the wall time. A job consists of both validation and testing, including the threshold fitting discussed in Section~\ref{sec:system}; only gemma-3-27b-pt reuses the thresholds already fitted for it and does not need to fit them again, which makes it the cheapest one. The most expensive is gemma-2-27b due to the inefficient attention implementation.

\begin{table}[h]
\centering\footnotesize
\setlength{\tabcolsep}{3pt}
\begin{tabular}{lcccc}
\toprule
& \multicolumn{2}{c}{probe} & \multicolumn{2}{c}{full system} \\
\cmidrule(lr){2-3}\cmidrule(lr){4-5}
checkpoint & wall & GPU-h & wall & GPU-h \\
\midrule
gemma-3-27b-pt & 0:03 & 0.1 & 0:21 & 0.7 \\
gemma-3-27b-it & 0:09 & 0.3 & 0:29 & 1.0 \\
gemma-2-27b & 0:04 & 0.1 & 0:38 & 1.3 \\
gemma-2-27b-it & 0:05 & 0.2 & 0:23 & 0.8 \\
gemma-3-12b-pt & 0:08 & 0.3 & 0:29 & 1.0 \\
gemma-3-12b-it & 0:12 & 0.4 & 0:22 & 0.7 \\
Mistral-Nemo-Base-2407 & 0:07 & 0.2 & 0:25 & 0.8 \\
Mistral-Nemo-Instruct-2407 & 0:10 & 0.3 & 0:24 & 0.8 \\
\midrule
Mistral-Small-3.2-24B & 0:06 & 0.2 & 0:33 & 1.1 \\
Magistral-Small-2506 & 0:05 & 0.2 & 0:27 & 0.9 \\
Qwen3-30B-A3B & 0:06 & 0.2 & 0:31 & 1.0 \\
Qwen3-30B-A3B-Think. & 1:01 & 2.0 & 0:28 & 0.9 \\
\bottomrule
\end{tabular}
\caption{Wall time and GPU-hours per checkpoint; two H100s per job. A full run
covers validation and test with fitting. Below the rule: no base sibling.}
\label{tab:compute}
\end{table}

\section{The Instruction-tuned Comparator in Full}
\label{app:chat}

This appendix outlines the instruction-tuned system deployed for the "it, chat" row in Table~\ref{tab:models}. Of all the instruction-tuned models we experimented with, this is our best system, which scored 0.5681 with gemma-3-27b-it and 0.6127 with Mistral-Small-3.2-24B. The second model achieved 0.6161 when the border rules discussed in Appendix~\ref{app:rules} were applied. This system utilizes a very simple process that entails generating the potential answers, evaluating them separately, calibrating the scores, and choosing the highest-scoring combination.
\paragraph{Propose.} The relation question is asked of the model for each row $n$ times. Instead of using temperature to generate different outputs, we provide each sample with a new \emph{approach instruction}, which is a one-sentence hint on how to solve the problem, e.g., "Walk around the country's borders clockwise and name its neighbours" or "First figure out whether the person was alive in 2026." In practice, it results in more productive disagreements among the samples compared to changing the temperature. Constrained decoding is used to make sure that the outputs remain in the same JSON format. Each of the prompts contains the definition of the relation given in the task and six example rows from the training set.

\paragraph{Group.} Identical to the main system: the scorer's own
normalisation for strings, 5\% tolerance clusters for numbers.

\paragraph{Verification.} Each candidate solution is then verified individually. The model receives a statement of the form "X is one of the correct answers to \ldots", and replies with either "Yes" or "No". Rather than sampling this reply, we take the log-probabilities of the first token: the score $q$ is the probability mass on the tokens spelling "Yes", divided by the mass on "Yes" plus "No". The value of $q$ obtained in this way is not always equally robust for different relations. For instance, it achieves an AUC of 0.931 for \rel{awardWonBy} while reaching 0.450 for the border relation, a result inferior even to the random chance baseline. Owing to such variations, we do not fix any threshold for $q$; rather, we consider it to be one signal among others forwarded to the next step.

\paragraph{Calibrate.} For each relation $r$, a logistic regression fitted using the training rows maps the signals to a probability. Let $f_i$ be the recurrence probability for candidate $i$, $q_i$ its verification probability, $\rho_i$ its rank among the row's candidates in $(0,1)$ space, and $\pi_i = 1/(1 + C)$ with $C$ being the number of candidates; this is a five-parameter model per relation (four weights and a bias), fitted with an $L_2$ penalty.

\paragraph{Decode.} The system outputs the set having the highest expected
row score based on these probabilities. Rank the candidates by $p_i$ and let
$S_k$ contain the top $k$; then only the $C{+}1$ sets $S_0, \dots, S_C$ have
to be compared \citep{dembczynski2011exact,ye2012optimizing}:
\begin{equation}
\label{eq:argmax}
S^{\ast} = \operatorname*{arg\,max}_{k \in \{0,\dots,C\}}
\; \mathbb{E}\!\left[F_1(S_k, G)\right].
\end{equation}
In the case when the real set can have at most one member, the candidates are mutually exclusive, and the expectation admits a closed-form solution; otherwise, they are assumed to be independent, and the expectation is computed using dynamic programming on the $p_i$. The null set is the special case of $k{=}0$ and therefore does not require any threshold. In the case of \rel{awardWonBy}, the model also infers the number of recipients and thus accounts for correct answers that are not in the candidate pool.

\paragraph{The Border Rules.}
\label{app:rules}
The border relation is characterized by two features which a purely statistical pipeline cannot take advantage of: the borders are symmetrical, and each training/validation instance is a full list of its subject's neighbours. Representing by $\mathcal{G}(X)$ the reference list of neighbours of a country $X$ that is a subject of the provided data, and by $\mathrm{dom}(\mathcal{G})$ the set of such subjects,
\begin{align}
\label{eq:closure}
Y \in \mathcal{G}(X) &\;\Rightarrow\; \textit{border}(Y, X),\\
\label{eq:cwa}
X \in \mathrm{dom}(\mathcal{G}) \wedge Y \notin \mathcal{G}(X)
&\;\Rightarrow\; \neg\,\textit{border}(Y, X).
\end{align}
Rule~\ref{eq:closure} adds missing neighbours (checked against the validation gold, the rule proposed 47 neighbours and 46 were correct; during test, on the Mistral-Small-3.2-24B chat system, it added five cases like The Gambia for Senegal); Rule~\ref{eq:cwa} deletes predicted neighbours which contradict the complete reference (during test, it deleted four cases like Sweden--Denmark, linked by a bridge not counted in the reference).

Both of them use reference data only, and predictions by the system itself are not considered evidence at all. In the case of the Mistral-Small-3.2-24B instruction-tuned system, the rules increased borders from 0.9487 to 0.9729 official and from 0.6127 to 0.6161 total. In the completion system, in which borders achieve 0.989 on validation with recall 1.000, the rules failed to pass the validation gate and were not used (Section~\ref{sec:results}).

\end{document}